\documentclass[a4paper]{article}
\usepackage{xcolor}
\newcommand{\panos}[1]{{\color{orange}#1}}
\renewcommand{\panos}[1]{}
\pdfoutput=1

\usepackage{INTERSPEECH2018}
\usepackage[none]{hyphenat}
\usepackage{amsmath,enumitem}
\usepackage{lipsum,graphicx,multicol}
\usepackage{cleveref}
\usepackage{subcaption}

\newcommand{\dq}[1]{``#1''}
\DeclareMathOperator*{\argmax}{arg\,max}

\graphicspath{ {figures/} }

\makeatletter
\newsavebox\myboxA
\newsavebox\myboxB
\newlength\mylenA
\newcommand*\xoverline[2][0.75]{%
    \sbox{\myboxA}{$\m@th#2$}%
    \setbox\myboxB\null
    \ht\myboxB=\ht\myboxA%
    \dp\myboxB=\dp\myboxA%
    \wd\myboxB=#1\wd\myboxA
    \sbox\myboxB{$\m@th\overline{\copy\myboxB}$}
    \setlength\mylenA{\the\wd\myboxA}
    \addtolength\mylenA{-\the\wd\myboxB}%
    \ifdim\wd\myboxB<\wd\myboxA%
       \rlap{\hskip 0.5\mylenA\usebox\myboxB}{\usebox\myboxA}%
    \else
        \hskip -0.5\mylenA\rlap{\usebox\myboxA}{\hskip 0.5\mylenA\usebox\myboxB}%
    \fi}
\makeatother

\title{Modeling Interpersonal Influence of Verbal Behavior\\in Couples Therapy Dyadic Interactions}
\name{Sandeep Nallan Chakravarthula$^1$, Brian Baucom$^2$, Panayiotis Georgiou$^1$}
\address{
  $^1$Dept. of Electrical Engineering, University of Southern California\\
  $^2$Dept. of Psychology, University of Utah}
\email{nallanch@usc.edu, brian.baucom@psych.utah.edu, georgiou@sipi.usc.edu}

\begin{document}

\maketitle
\begin{abstract}
  Dyadic interactions among humans are marked by speakers continuously influencing and reacting to each other in terms of responses and behaviors, among others. Understanding how interpersonal dynamics affect behavior is important for successful treatment in psychotherapy domains. Traditional schemes that automatically identify behavior for this purpose have often looked at only the target speaker. In this work, we propose a Markov model of how a target speaker's behavior is influenced by their own past behavior as well as their perception of their partner's behavior, based on lexical features. Apart from incorporating additional potentially useful information, our model can also control the degree to which the partner affects the target speaker. We evaluate our proposed model on the task of classifying Negative behavior in Couples Therapy and show that it is more accurate than the single-speaker model. Furthermore, we investigate the degree to which the optimal influence relates to how well a couple does on the long-term, via relating to relationship outcomes.
\end{abstract}
\vspace{-0.25cm}

\section{Introduction}
\label{sec:intro}

Conversations in social settings are often marked by each person expressing behaviors that are not only driven by their own internal state of mind but also affected by how the other person responds to them \cite{mikulincer2002attachment}. The nature of this phenomenon, referred to as interpersonal influence, can vary significantly, depending on the individual traits of the speakers as well as the relationship between them. Even for the same speaker, the type of influence might be different with different people, causing their behavior to change differently. For example, a person might respond positively when their friend compliments them but might respond in a lukewarm manner when a stranger compliments them. Therefore, creating models that explicitly incorporate influence can provide new understanding of the longer-term dyad dynamics along with better behavioral estimation.

\panos{Need to be careful: entrainment is a dynamic/2-person feature; Matt's work included both interlocutors in model but lacked dynamics. Not all psychotherapy is couples... Tone claims and add non USC citations too....}
Observational studies in psychotherapy domains involve interactions that are observed and assessed by human annotators on certain behaviors of interest.
In Behavioral Signal Processing \cite{georgiou2011_behavioral-sign, narayanan2013_behavioral-sign} we employ automated scoring of these interactions using speech and language analysis with machine learning to quantify these same behaviors. There are multiple methods of doing this, but one very successful technique has been through analysis of the spoken language of the rated speaker.  Most work in this regard, however, either did not explicitly model the underlying states of the interlocutors \cite{black2010automatic,tseng2016couples} or completely ignored the rated speaker's partner \cite{chakravarthula2015language,xiao2015_rate-my-therapi}. These models lack explicit understanding of the interpersonal influence. While interlocutor influence models have been proposed in the past \cite{lee2009modeling,yang2015modeling}, they assume direct interaction between the latent states of the speakers which is not applicable for our problem. Outside of BSP, mutual influence models have been proposed, such as \cite{cook2005actor} which predicts \dq{attachment security} between mother and child based on their previous measures of security. However, these deal with fully observed processes which are different from the latent behavior we are interested in modeling.


Our proposed model, which we refer to as the influence model, describes how a speaker continuously perceives their partner's behavior based on their responses and how this perception, in addition to their own past behavior, affects their behavior over time. It also specifies parameters that determine the strength of the influence mechanisms. Thus, it provides a more complete understanding of how and why a person's behavior changes over time.

Additionally, we investigate if the interpersonal influence between the speakers of a couple during therapy relates to their relationship outcomes. It has been shown that interlocutor influence in couples dyadic interactions is important for relationship functioning \cite{baucom2015changes}, and therefore, outcomes. Empirical relations have also been found between outcomes and interaction-dependent measures such as vocal entrainment (through withdrawal behavior) \cite{lee2012using} and dyadic prosodic features \cite{nasir2017predicting}. Therefore, we want to examine whether our model parameters, that are designed to describe the characteristics of dyadic interactions, can also provide information about outcomes.


\section{Behavior Interaction Model}
\label{sec:model}

We first present our baseline model, proposed in \cite{chakravarthula2015language}, in Sec.~\ref{ssec:ldbm} and then present our proposed influence model in Sec.~\ref{ssec:inf}


\subsection{Likelihood-based Dynamic Behavior Model (LDBM)}
\label{ssec:ldbm}

The LDBM is a single-speaker model that characterizes a person expressing a certain \emph{class} of behavior (ex: \dq{High Anger} or \dq{Low Anger}) as an HMM-based generative process.
Specifically, in each turn, the target speaker occupies a latent behavior \emph{state}, generates an utterance with some emission probability, either remains in the same state or moves to a different one with some transition probability, generates a new utterance and repeats this until the end of the interaction.
The full behavior process is given by:
\vspace{-0.25cm}

\begin{flalign}
P(\bm{S_{s}},\bm{U_{s}})=P(\bm{S_{s}})P(\bm{U_{s}}|\bm{S_{s}})& \label{eq:ldbm_joint}
\end{flalign}

\begin{figure}[!t]
    \includegraphics[width=1\linewidth]{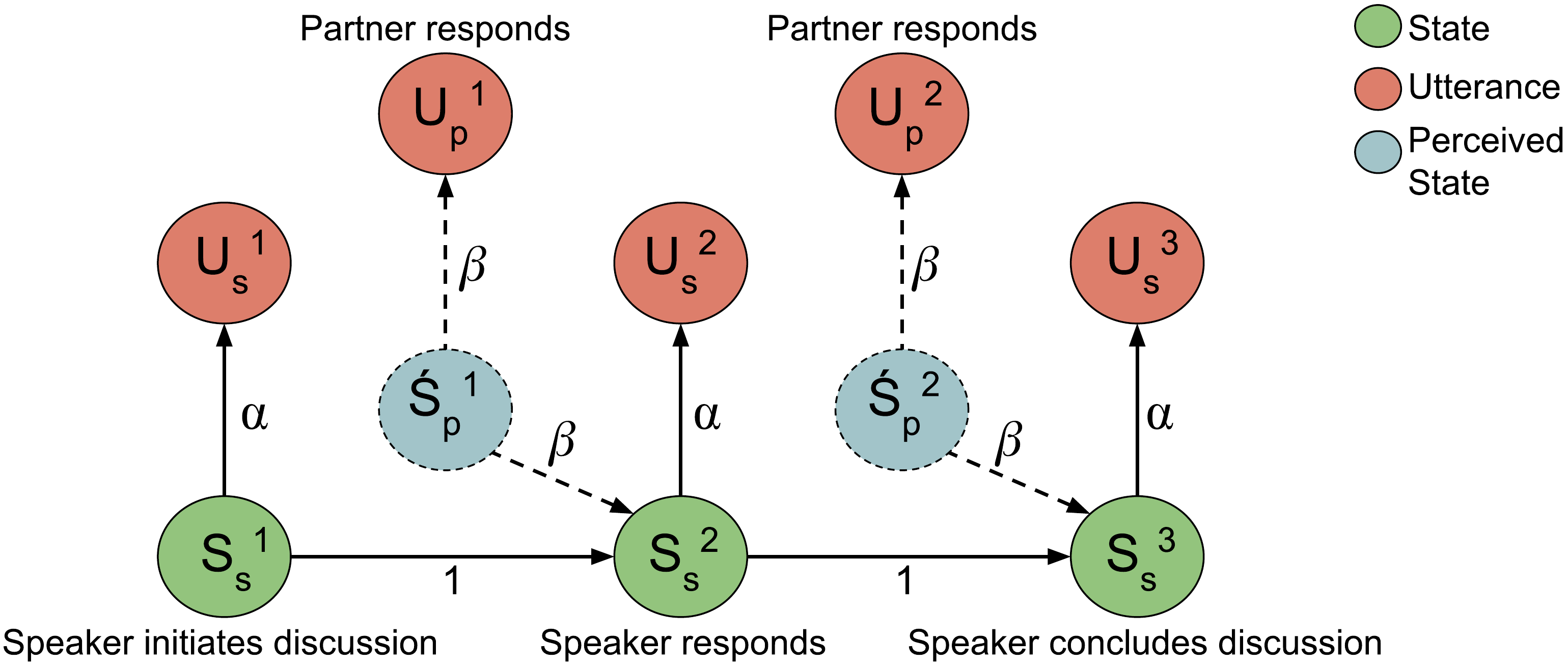}
    \caption{\it{Behavior generation process over 3 speaker turns in the Influence Model}}
    \label{fig:new_model}
\vspace{-0.5cm}
\end{figure}

where $\textbf{S}_{s}$ denotes the speaker states and $\textbf{U}_{s}$ denotes the utterances generated while in those states. Behavior classes in the LDBM share the same set of states but differ in their transition probabilities. For example, a \dq{High Anger} speaker might tend to stay in a particular state and not change whereas a \dq{Low Anger} speaker might frequently change states. The LDBM is the current best model in behavioral classification as it incorporates a more accurate model of time-varying behavior than then-existing methods. It does not however  take the partner's effect into account.
\vspace{-0.35cm}

\subsection{Influence Model}
\label{ssec:inf}

As an extension, our proposed model introduces an additional latent \dq{pseudo-state} that represents how the target speaker perceives their partner's behavior. Its purpose is to simulate the generation of the partner's utterance and affect the next state of the speaker. We do this to model the real-world scenario where a speaker does not know the state of mind of the partner they are interacting with \cite{mikulincer2002attachment} - the best they can do is to make an inference based on what they said. We call the partner's \emph{perceived-by-target-speaker} state a pseudo-state to distinguish it from a state which represents the partner's internal state of mind. We denote the target speaker, the one whose behavior we are interested in identifying, by $S_{s}$ and the partner, from whom we are obtaining supplementary information, by $\acute{S}_{p}$.

Furthermore, our influence model specifies parameters that control how much a speaker's behavior is influenced by their own past behavior versus their partner, motivated by the domain literature \cite{berns1999demand}. $\alpha$ and $\beta$ respectively control how much the speaker's current utterance and partner explain their current behavior, relative to their previous state, which is always weighted by 1. Speakers who do not listen much to their partner and instead continually espouse only their point of view have $\beta \ll \alpha,1$. In contrast, speakers that strongly react to what their partners say during discussions have $\beta \geqslant \alpha,1$.

The behavior generation process is as follows: In a speaking turn $i$, the target speaker, who is currently in state $S_{s}^{i-1}$, observes the partner generated utterance $U_{p}^{i-1}$ and perceives their partner to be in pseudo-state $\acute{S}_{p}^{i-1}$. Then, based on $S_{s}^{i-1}$ and $\acute{S}_{p}^{i-1}$, the speaker transitions to state $S_{s}^i$ and generates $U_{s}^i$. As an illustration, Fig.~\ref{fig:new_model} depicts this process over 3 turns of an interaction.
In this model, while $S_{s}^i$ depends on $S_{s}^{i-1}$, $\acute{S}_{p}^i$ does not depend on $\acute{S}_{p}^{i-1}$.

For an interaction consisting of $M$ turns, the complete behavior generation process is given by:

\begin{eqnarray}
\label{eq:class_dec_cond}
  \begin{split}
    P(\bm{S_{s}},\bm{\acute{S}_{p}},& \bm{U_{s}},\bm{U_{p}};\alpha,\beta) =  \\
= & P(\bm{S_{s}},\bm{\acute{S}_{p}};\beta)P(\bm{U_{s}},\bm{U_{p}}|\bm{S_{s}},\bm{\acute{S}_{p}};\alpha,\beta)  \\
= & P(S_{s}^1) \cdot \prod\limits_{j=2}^M P(S_{s}^j|S_{s}^{j-1}) \cdot \prod\limits_{j=2}^{M-1} P(S_{s}^j|\acute{S}_{p}^{j-1})^\beta \\
& \ \ \ \ \ \ \ \ \ \ \ \cdot \prod\limits_{j=1}^M P(U_{s}^j|S_{s}^j)^\alpha \cdot \prod\limits_{j=1}^{M-1} P(U_{p}^j|\acute{S}_{p}^j)^\beta \\
\end{split}
\end{eqnarray}

\noindent where \textbf{A} denotes the set $\{A^1,\ldots,A^M\}$ for target speaker and $\{A^1,\ldots,A^{M-1}\}$ for partner. $P(S_{s}^j|S_{s}^{j-1})$ and $P(S_{s}^j|\acute{S}_{p}^{j-1})$ are transition probabilities that denote how a speaker's previous state and their partner's previous pseudo-state are likely to affect the speaker's choice of next state. $P(U_{s}^j|S_{s}^j)$ is an emission probability describing the likelihood of the speaker saying something when occupying that state. $P(U_{p}^j|\acute{S}_{p}^j)$ describes the likelihood, according to the speaker, of the partner saying something in the perceived state. Since both emission probabilities deal with the speaker's perspective they are obtained from the same model.
\vspace{-0.125cm}

\section{Data}
\label{sec:data}

The Couples Therapy corpus \cite{christensen2004} contains audio, video recordings and manual transcriptions of conversations between 134 real-life couples attending marital therapy. In each session, one person selected a topic that was discussed over 10 minutes with the spouse. At the end of the session, both speakers were rated separately on 33 \dq{behavior codes} by multiple annotators based on the Couples Interaction \cite{heavey2002couples} and Social Support  \cite{Jones1998} Rating Systems. Each behavior was rated on a Likert scale from 1, indicating absence, to 9, indicating strong presence. A session-level rating was obtained for each speaker by averaging the annotator ratings. This process was repeated for the spouse, resulting in 2 sessions per couple at a time. The total number of sessions per couple varied between 2 and 6. Henceforth, for couples, we refer to the speaker whose behavior we want to classify as \dq{rated speaker} and their partner as \dq{spouse}.

For classifying Negative behavior, we used the binarized dataset where sessions with ratings in the top 20 percentile i.e. High Negative are assigned the class label $C_1$ while those in the bottom 20 percentile i.e. Low Negative are assigned $C_0$. There are 280 sessions in total with 140 in each class. In each session, we formatted the interaction to enforce the rated speaker to begin the conversation as well as end it as illustrated in Fig.~\ref{fig:new_model}.


Each couple was also rated at the 26-week or the 2-year time period on its recovery since starting therapy, referred to as \dq{outcomes} \cite{baucom2011observed}. They were rated as follows - 1 indicating deterioration in relationship, 2 indicating no change, 3 indicating measurably better improvement and 4 indicating recovery. We use 2-year outcome ratings for 79 couples belonging to the Top/Bottom 20 percentile classes. These are the ones we are interesting in analyzing and their demographics are shown in Table \ref{tab:dem_out}.

\begin{table}[!htbp]
\centering
\begin{tabular}{*5c}
\toprule
Outcome & Decline & No Change & Better & Recovery\\
\midrule
Rating & 1 & 2 & 3 & 4\\
No. Couples & 18 & 9 & 18 & 34 \\
\bottomrule
\end{tabular}
\caption{Couples Therapy Outcome Demographics}
\label{tab:dem_out}
\vspace{-0.875cm}
\end{table}

\section{Experiments}
\label{sec:exp}

In this section, we describe the classification experiment. We also describe in Sec.~\ref{ssec:exp_out} how we can employ the optimized model parameters, $\alpha, \beta$, as indicators of influence or inertia and how these can be related to  outcomes.\vspace{-0.25cm}

\subsection{Behavior Classification}
\label{ssec:exp_class}


We use a full leave-one-couple-out train-dev-test scheme in order to account for the small number of samples and to prevent over-fitting to the speaking styles or interaction topics of the test couple. Model parameters $\alpha$ and $\beta$ are tuned on the dev set. We evaluate our model against the LDBM with 1-gram, 2-gram and 3-gram LMs to examine the effect of longer context. We also modify the LDBM process described in Eqn.~\ref{eq:ldbm_joint} to allow scaling of the emission probabilities with the same $\alpha$ used in our model. This allows for a fairer comparison of the two models.\vspace{-0.25cm}


\subsubsection{Training}
\label{sssec:class_train}
\vspace{-0.125cm}
The core training methodology is similar to that of the LDBM in our previous work \cite{chakravarthula2015language}. We use two states $S_0$ and $S_1$ which are represented by statistical language models that we term language-to-behavior (L2B) models. In this case L2B employs an ARPA format as in \cite{georgiou2011thats-aggravati,chakravarthula2015language}. Both classes $C_0$ and $C_1$ use the same states but through different state transition probabilities. The state models, along with the transition probabilities, are trained using the Expectation Maximization (EM) algorithm.

In this first attempt at creating the influence model, our focus was on establishing its benefits. Thus, rather than risk overfitting through optimization, we decided to explore a range of $\alpha$ and $\beta$ in logarithmic steps of 10 from $10^{-3}$ to $10^3$, resulting in a search space of 49 points. The main idea is to get a general sense of the importance of the information streams. For example, $\alpha$$=$$1$, $\beta$$=$$1$ implies that \dq{the rated speaker is affected as much by their own past behavior as their spouse's}.


The training procedure is as follows:

\begin{enumerate}[itemsep=-0.5mm]
\item In a test fold, pick parameter configuration $\alpha,\beta$
\item Initialize and iteratively train state models with $\alpha,\beta$ using the LDBM training described in \cite{chakravarthula2015assessing}
\item Estimate state transition probabilities per class, classify dev couple sessions and repeat for all dev sessions
\item Pick best configuration $\alpha,\beta$ based on dev accuracy
\item Use this $\alpha, \beta$, train class and state models to evaluate on the test session and repeat for all test folds
\end{enumerate}

We avoided estimating the transition probabilities at each iteration since they tended to skew towards not changing states as a result of the turns initially having the same label. Also, at the end of the training stage, we would normally end up with only one model. In this case, though, we have a set of models, one for each test couple, because of the specific train-dev-test setup used. The training procedure for the modified LDBM is similar with only $\alpha$ being optimized over the dev set. For baseline comparison, we also trained LDBM with fixed $\alpha=1$
\vspace{-0.25cm}

\subsubsection{Testing and Evaluation}
\label{sssec:class_test}
Given a session consisting of $M$ rated speaker turns $\textbf{U}_{s}$ and $M$$-$$1$ spouse turns $\textbf{U}_{p}$, the goal is to classify the rated speaker as either $C_0$ or $C_1$. For each class, we get decoded state sequences, that best explain the observed utterances, along with its probability. The class that is most likely to have generated the utterances is then picked as the label of the rated speaker as denoted below:
\begin{eqnarray}
C_i&=&\argmax\limits_{C_j}P(\bm{S_{s}^j},\bm{\acute{S}_{p}^j}|\bm{U_{s}}\bm{U_{p}};\alpha,\beta)\label{eq:class_dec} \\ 
&=&\argmax\limits_{C_j}P(\bm{S_{s}^j},\bm{\acute{S}_{p}^j},\bm{U_{s}},\bm{U_{p}};\alpha,\beta) \label{eq:class_dec_joint}
\end{eqnarray}

where 
$\bm{S_{s}}^j$ is the speaker state sequence decoded by model of $C_j$ for $\bm{U_{s}}$, 
$\bm{\acute{S}_{p}^j}$ is the perceived partner state sequence decoded by model of $C_j$ for $\bm{U_{p}}$, 
and $\alpha$ and $\beta$ are the optimized parameters used to train final model

The joint probability in Eqn.~\ref{eq:class_dec_joint} reduces to the conditionally independent probabilities in Eqn.~\ref{eq:class_dec_cond}. Using this scheme, we classify all test couples with their corresponding fold models trained in Sec.~\ref{sssec:class_train} and compute the classification accuracy. The same scheme is also used to test the LDBM and obtain its classification accuracy. The two models, along with the baseline, are then compared in Sec.~\ref{ssec:res_class}.\vspace{-0.25cm}

\subsection{Mapping Parameters to Outcomes}
\label{ssec:exp_out}

We investigated whether our model parameters contain information about couples outcomes using the following procedure: For an outcome rating, select couples with this rating. Select a couple and obtain its average dev classification accuracy over all test folds for each parameter configuration $\alpha$, $\beta$. Pick the parameter configuration with the highest accuracy and add 1 count to its bin; equal fractional counts if there are multiple configurations. Repeat for all couples and normalize the counts to get a 2-D histogram. Convert 2-D histogram into a 1-D histogram with new axis $\gamma=log10(1/\beta)$ and repeat for all outcomes. The results of this analysis are discussed in Sec.~\ref{ssec:res_out}.\vspace{-0.125cm}

\section{Results \& Discussion}
\label{sec:results}

\subsection{Behavior Classification}
\label{ssec:res_class}

The test results comparison betwen the baseline LDBM ($\alpha$$=$$1$), LDBM and influence model is shown in Table.~\ref{tab:res_class}.

\begin{table}[ht]
\centering
\begin{tabular}{*4c}
\toprule
Model & LDBM ($\alpha$$=$$1$) & LDBM & Influence Model\\
\midrule
1-gram & 84.64 & 82.86 & \textbf{85.00}\\
2-gram & 86.78 & 86.43 & \textbf{88.93}\\
3-gram & 85.71 & 87.86 & \textbf{88.21}\\
\bottomrule
\end{tabular}
\caption{Comparison of Test Classification Accuracy \% with best performing model indicated in \textbf{bold}}
\label{tab:res_class}
\vspace{-0.5cm}
\end{table}

\begin{figure*}
    \centering
    \begin{subfigure}[b]{0.35\textwidth}
        \includegraphics[width=1.1\textwidth]{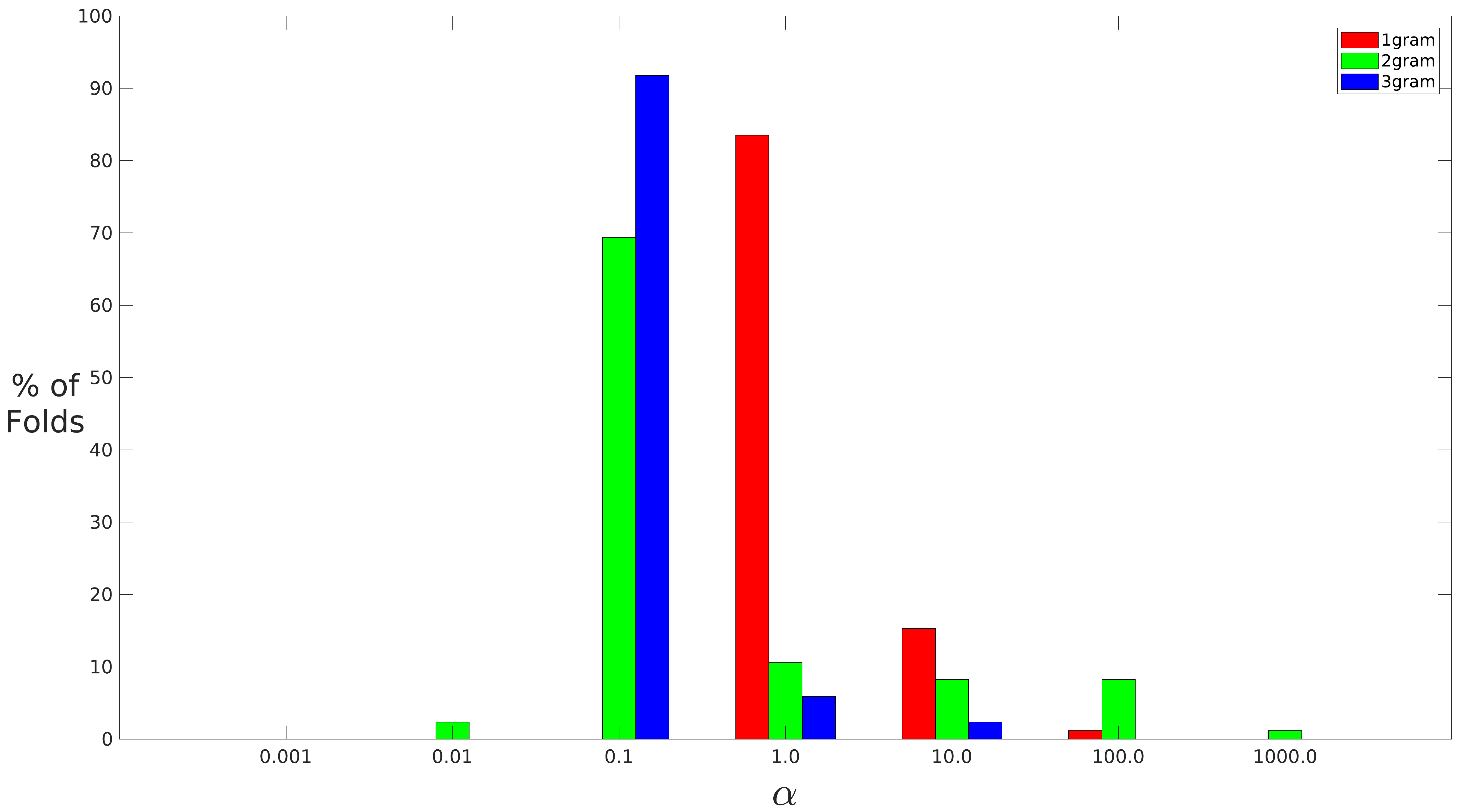}
        \caption{{LDBM 1,2,3-gram}}
        \label{fig:old}
    \end{subfigure}
	\	\	\	\	\	\	\	\	\	\	\	\	\	\	\	\	\	\	\	\	\	\	\	\	\	\	\	\
    \begin{subfigure}[b]{0.35\textwidth}
        \includegraphics[width=1.1\textwidth]{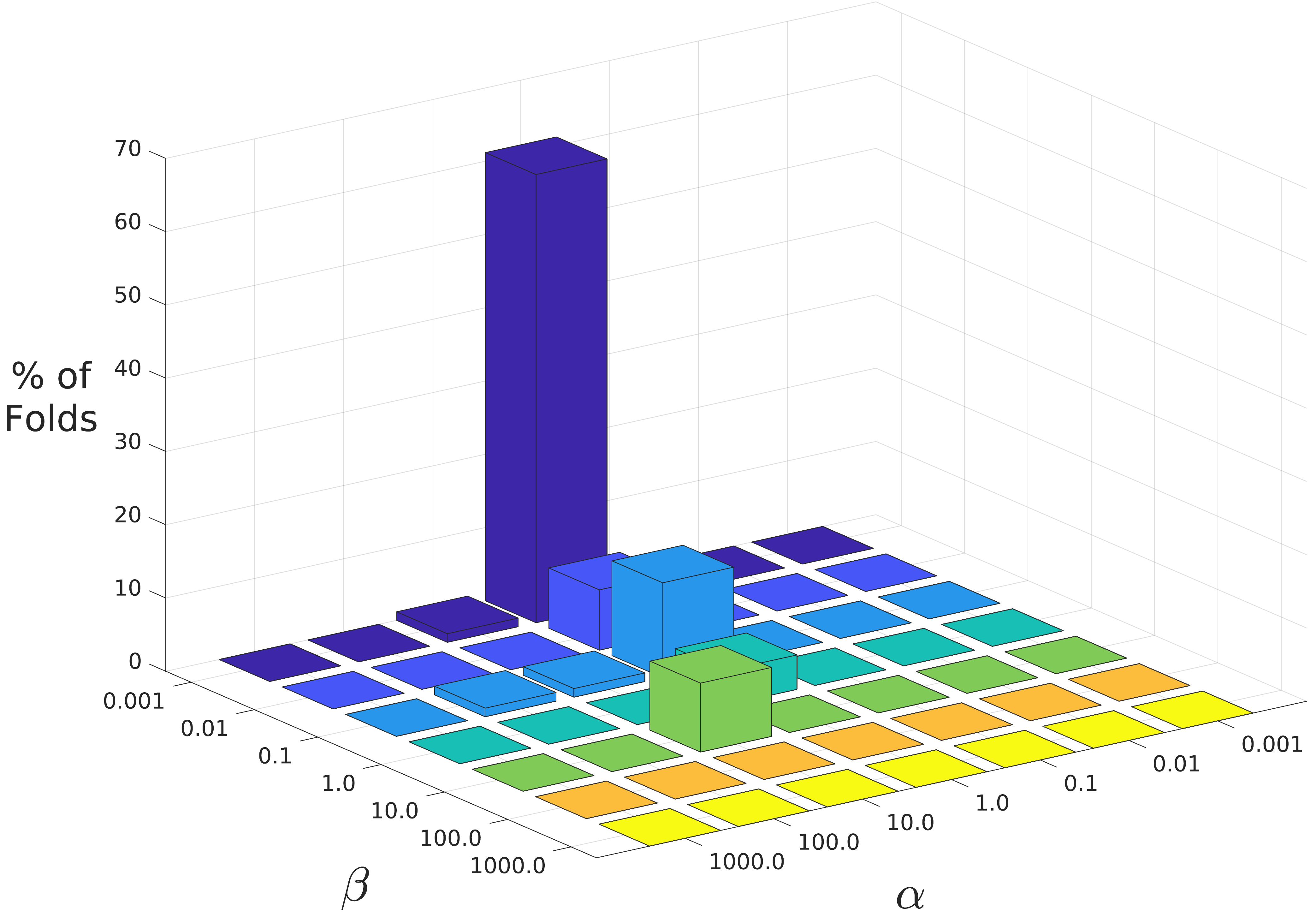}
        \caption{{Influence 1-gram}}
        \label{fig:new1}
    \end{subfigure}
    \begin{subfigure}[b]{0.35\textwidth}
        \includegraphics[width=1.1\textwidth]{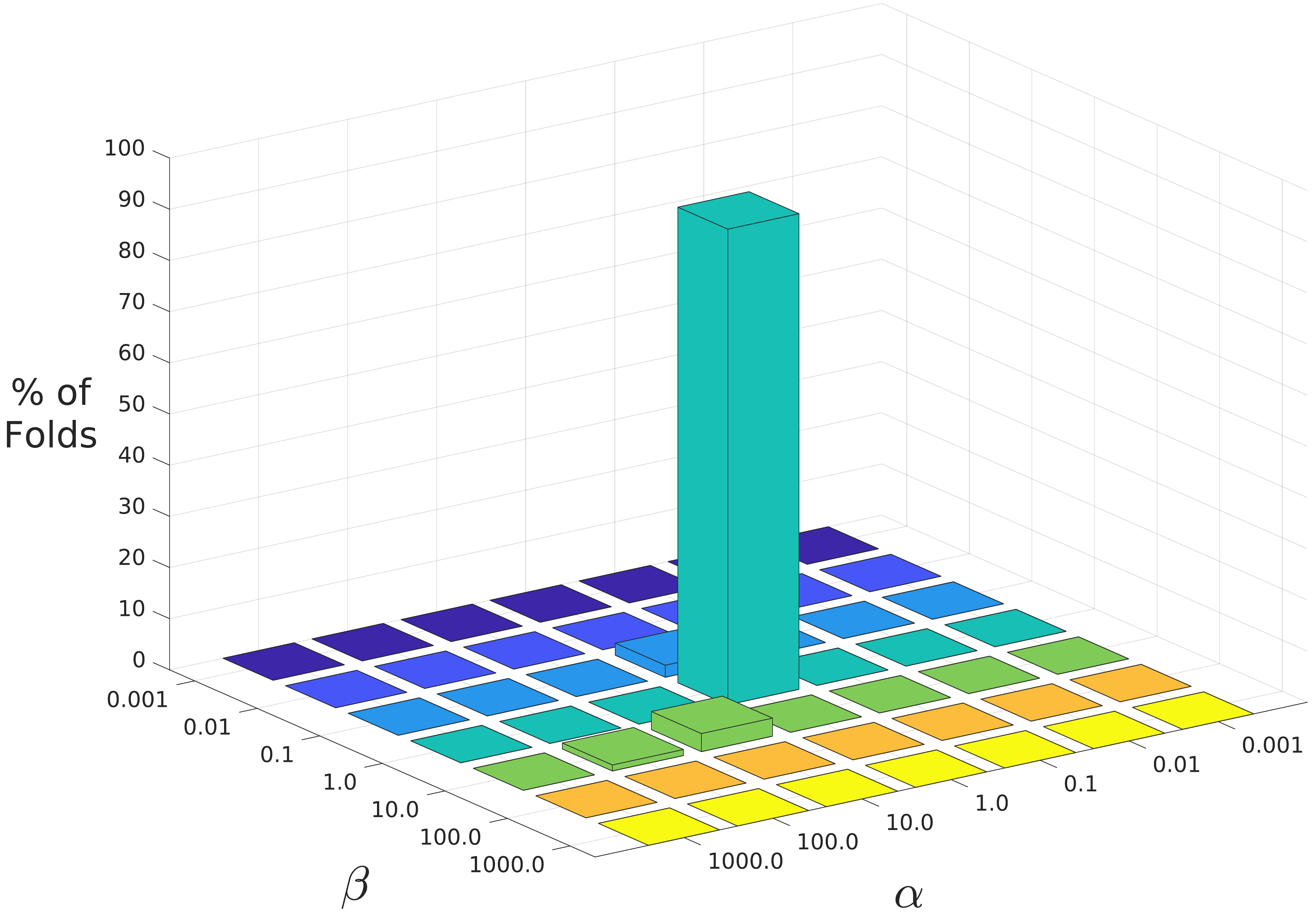}
        \caption{{Influence 2-gram}}
        \label{fig:new2}
    \end{subfigure}
	\	\	\	\	\	\	\	\	\	\	\	\	\	\	\	\	\	\	\	\	\	\	\	\	\	\	\	\
    \begin{subfigure}[b]{0.35\textwidth}
        \includegraphics[width=1.1\textwidth]{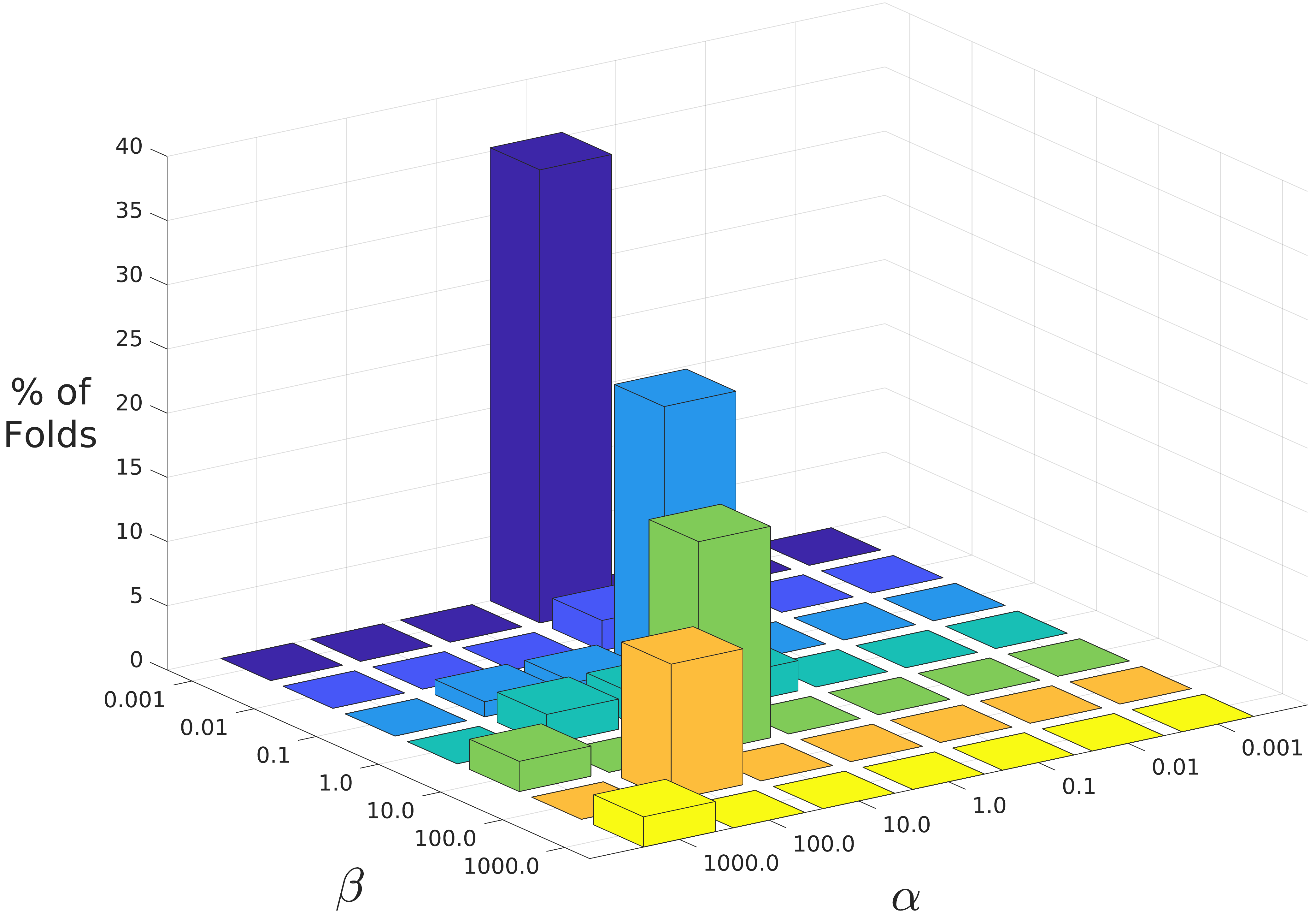}
        \caption{{Influence 3-gram}}
        \label{fig:new3}
    \end{subfigure}
    \caption{Histogram of best parameter configurations for 1,2,3-gram LDBM and Influence models}\label{fig:visualization}
\end{figure*}

We see that our proposed model improves upon the existing one and the baseline in every context scenario. This indicates that the partner's responses do provide supplementary information about why the speaker behaves in a certain way. We also see that while the LDBM gets better with more context, the influence model improves from 1-gram to 2-gram but slightly drops in accuracy from 2-gram to 3-gram. One reason for this could be the coarse sampling of control parameters while another could simply be the sparsity in learning 3-grams in bigger models. We note that our best performing model matches the best RNN-based accuracy obtained by Tseng et al in \cite{tseng2016couples}. This points to complementary gains from improved L2B models \cite{tseng2016couples} and from incorporating interlocutor dynamics as in this work.

We also examined which $(\alpha,\beta)$ configuration in the influence model and $\alpha$ in the LDBM performed best in different test folds. This can provide insight into how relevant each stream of information is, on average. The histogram of best configurations for both models is shown in \cref{fig:old,fig:new1,fig:new2,fig:new3}. It is interesting to note that the best performing model, the 2-gram influence model, overwhelmingly picked $\alpha$$=$$\beta$$=$$1$, implying equal importance of all the information streams. This too stresses the importance of incorporating the spouse's feedback when studying a rated speaker's Negative behavior.\vspace{-0.125cm}

\subsection{Mapping Parameters to Outcomes}
\label{ssec:res_out}


The 1-D histogram relates to interpersonal influence in the following way: $\gamma=0$ denotes equal contribution of the rated speaker's past behavior as the spouse.
$\gamma>0$ corresponds to the speaker's past behavior being more dominant than the spouse whereas $\gamma<0$ denotes the reverse.

In order to check if the outcome histograms are generated from different distributions, we fit a Gaussian distribution to each of them.
The resulting distributions have high variance and the following mean $\mu$ values: 0.20, 0.16, -0.18 and 0.05 for outcomes 1,2,3 and 4 respectively.
We see that the worst outcome, 1, has its $\mu$ farthest from $0$ while the best outcome, 4, has its $\mu$ closest to $0$. Thus, we see a link between the best/worst outcomes and the model parameters. These are inconclusive results but they are encouraging enough to warrant further investigation in future.
    \vspace{-0.35cm}

\section{Conclusions \& Future Work}
\label{sec:confut}

In this work, we proposed a model of how a speaker's behavior changes over time as a result of interacting with their partner. Our approach described a generative process of how one person is perceived based on what they said and how this perception relatively affects the other in conjunction with their own past behavior. By achieving higher accuracy over a single-speaker model on the task of classifying Negative behavior for Couples Therapy, we demonstrated the effectiveness of incorporating information from both participants in a conversation.

In addition, we investigated if our model could provide insights into how interaction dynamics relate to the long-term quality of a couple's relationship. We found that some outcomes tend to be associated with how proportionately a speaker is affected by their partner's perceived behavior with respect to their own past behavior.

As part of future work, we will implement comprehensive optimization of the model while also jointly training all components in it. We also plan on investigating which behaviors benefit from influence modeling and which ones don't and why. Another area where our model can potentially be improved is in replacing discrete behavior states with continuous ones such as in the Kalman filter or recursive neural networks. We will also investigate incorporating our more recent advances in language-to-behavior mapping as in \cite{tseng2016couples}.

We also plan on further investigating the relation between influence parameters and outcomes with the help of the extensions mentioned above - refined estimation of parameters and for all behaviors - as well as by jointly analyzing the influence parameters and states.
    \vspace{-0.25cm}


\section{Acknowledgments}
\footnotesize
The U.S. Army Medical Research Acquisition Activity, 820 Chandler Street, Fort Detrick MD 21702-5014 is the awarding and administering acquisition office. This work was supported by the Office of the Assistant Secretary of Defense for Health Affairs through the Psychological Health and Traumatic Brain Injury Research Program under Award No. W81XWH-15-1-0632. Opinions, interpretations, conclusions and recommendations are those of the author and are not necessarily endorsed by the Department of Defense

\newpage

\bibliographystyle{IEEEtran}

\bibliography{mybib}

\end{document}